\documentclass[lettersize,journal]{IEEEtran}
\usepackage{amsmath,amsfonts}
\usepackage{amsthm}
\usepackage{array}
\usepackage{textcomp}
\usepackage{url}
\usepackage{graphicx}
\usepackage{cite}
\usepackage{pifont}
\usepackage{multirow}
\usepackage{makecell}
\usepackage{etoolbox}
\usepackage{xcolor}
\usepackage{booktabs}
\newbool{showcomments}
\setbool{showcomments}{false}
\newcommand{\ewa}[1]{\ifbool{showcomments}{\textcolor{blue}{[E: #1]}}{}}
\newcommand{\solved}[1]{\ifbool{showcomments}{\textcolor{green}{[Solved: #1]}}{}}
\newcommand{\gautier}[1]{\ifbool{showcomments}{\textcolor{red}{[G: #1]}}{}}
\definecolor{asrblue}{HTML}{185FA5}
\definecolor{tprorange}{HTML}{D85A30}
\newcommand{\ci}[1]{{\scriptsize $\pm$#1}}

\begin{document}

\title{The Forensic Cost of Watermark Removal: 
From Dedicated Attacks to Image Editing}

\author{Gautier Evennou$^{1,2}$, Ewa Kijak$^2$ \\
$^1$IMATAG, $^2$IRISA, Univ. Rennes, INRIA, CNRS 
}

\maketitle

\begin{abstract}
Current watermark removal methods are evaluated on two axes: attack success rate and perceptual quality. We show that this is insufficient. While state-of-the-art attacks successfully degrade the watermark signal without visible distortion, they leave distinct statistical artifacts that betray the removal attempt. We name this overlooked axis Watermark Removal Detection (WRD) and demonstrate that a modern classifier trained on these artifacts achieves state-of-the-art detection rates with an average of 0.92 TPR at $10^{-3}$ FPR across every removal method tested. We introduce the compound success rate to better report attacks that both erase the watermark and avoid forensic detection, revealing that the best practical threat achieves only 10\% of success rate. We provide extensive evaluation across text-to-image and editing models spanning 2022–2026, finding that GenAI faces a fundamental dilemma: either they fail to remove the watermark, or they introduce distortions strong enough to be forensically exposed. No current method balances all three criteria simultaneously.
\end{abstract}


\section{Introduction}
\IEEEPARstart{W}atermarking is under attack. As hidden signals become
the default mechanism for tracing image provenance, mandated by
regulators~\cite{euaiact2024} and adopted by
platforms~\cite{castro2025watermarking}, adversaries have strong
incentives to strip them. A successful removal enables
intellectual-property laundering, bypasses model-collapse safeguards
for GenAI, and severs chains of trust in information networks.

\begin{figure}
    \centering
    \includegraphics[width=\linewidth]{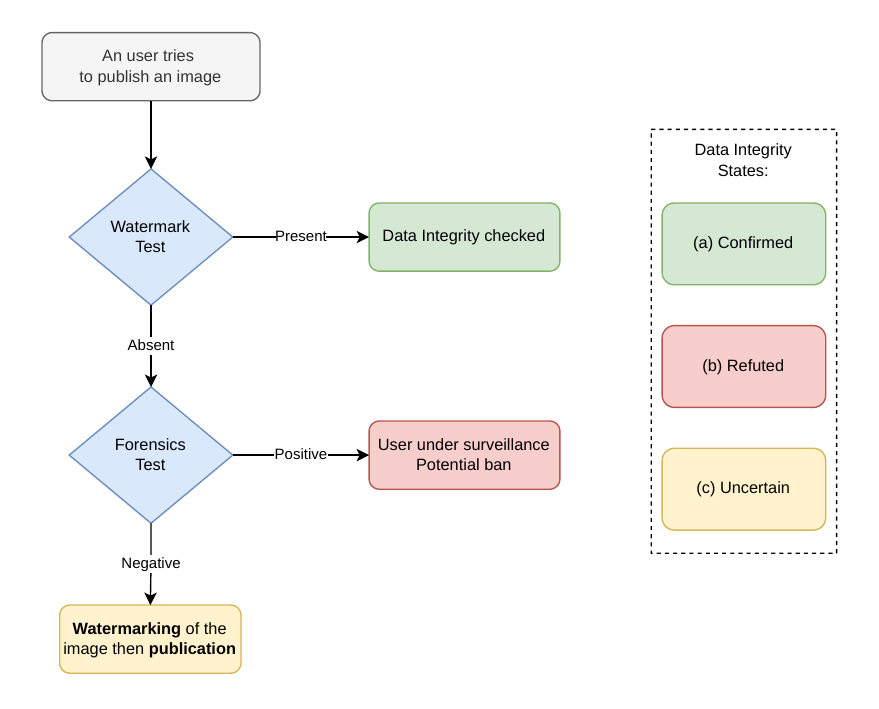}
    \caption{Workflow of our method. We adopt a social network perspective, in which a user attempts to publish an image. The platform performs two sequential checks; the watermark test determines whether a watermark is detected. If so, the extracted ID is used to establish data provenance and enable the cover verification. A negative result triggers the forensics test, which detects watermark removal. A positive result prompts further scrutiny by the social network, potentially leading to user ban. A negative result leaves the image's status uncertain, after which it is watermarked and finally published.}
    \label{fig:enter-label}
\end{figure}

To understand what a successful attack requires, consider the following
threat model. \textit{Alice} produces an image and publishes it online.
\textit{Bob}, a platform verifier, runs two sequential checks: first
for Alice's watermark, then, if the watermark is absent, for traces of
removal. \textit{Eve} copies the image and attempts removal to repost
it as her own. Bob checks Eve's image, detects no watermark but finds
removal artifacts, and triggers sanctions. A successful attack must
therefore fool \emph{both} detectors: it must remove the watermark
\emph{and} leave no forensic trace, as illustrated in Figure~\ref{fig:enter-label}. We assume Bob targets high-profile
users to limit top-down propagation of stolen media.

Yet current removal methods address only half of this requirement.
Whether adversarial or generative, they frame the task as noise
purification: the watermark is treated as additive noise, and success
is measured solely by the trade-off between Attack Success Rate (ASR)
and distortion (PSNR). If the watermark is unreadable and the
distortion low, the attack is deemed successful. This framing ignores
a critical question: \emph{does the post-removal image still look like
a natural image to a forensic detector?}

Our answer is no. We show that the act of removal leaves distinct
statistical trails, high-frequency adversarial noise, characteristic
diffusion artifacts or both, that push the image off the manifold of
natural images. These traces are readily separable by a learned
classifier trained on genuine non-watermarked images and
post-removal images, which flags attacks with high TPR at low FPR
across a wide range of state-of-the-art removal methods. To our
knowledge, no existing attack accounts for this forensic leakage.

We formalize this observation into a new evaluation axis:
\textbf{Watermark Removal Detection (WRD)}. Any complete assessment
of watermark removal must jointly consider three criteria:
(i)~attack success rate, (ii)~perceptual quality, and
(iii)~forensic stealthiness. Under this extended framework, we
demonstrate that all current removal methods fail criterion~(iii),
revealing stealth as a fatal gap in existing attack strategies.

Our contributions are:
\begin{itemize}
    \item We introduce Watermark Removal Detection as a novel
    evaluation constraint, formalizing forensic stealthiness as a
    necessary requirement for watermark removal.
    \item We show that a lightweight classifier reliably detects
    post-removal images across multiple removal families (adversarial,
    diffusion-based), and benchmark state-of-the-art methods under
    WRD, finding that none achieves forensic stealth.
    \item We extend the WRD evaluation to modern image editing models,
    finding that editors either fail to remove the watermark or
    introduce distortions strong enough to be forensically detected,
    establishing a fundamental dilemma for editing-based attacks.
\end{itemize}

This work is an extended version of our preliminary work accepted at the 2026 ACM Workshop on Information Hiding and Multimedia Security (IH\&MMSec) \cite{evennou2026forensiccostwatermarkremoval}. We introduce the compound success rate (Eq.~2), a single metric capturing the joint requirement of watermark removal and forensic stealth. We add a cross-watermark generalization study (Sec.~IV-D) validating that the forensic detector learns attack artifacts rather than watermark-specific signals, and a comparison against AEROBLADE~\cite{AEROBLADE} (Sec. \ref{sec:genai_detection}) showing that off-the-shelf GenAI detectors do not transfer to the watermark removal setting. The main addition is a full evaluation of image editing as an attack vector (Sec.~IV-I): we benchmark eight editors spanning 2022–2026, uncover a temporal trend where watermark removal as a byproduct peaks in 2024 and declines afterward, and explain it through a spectral shift from low- to mid/high-frequency edits. Editors that do remove the watermark introduce distortions strong enough to be forensically detected, placing image editing under the same dilemma as dedicated removal attacks.

\section{Related Work}

\paragraph{Image Watermarking}
Two paradigms coexist in watermarking: in-gen for GenAI models, which embed the signal during the generation process \cite{stablesignature,treerings}, and post-hoc, which rather watermark the cover after its production, whether human or artificial. In this study we focus on post-hoc watermarking. 
Post-hoc image watermarking embeds a signal into an existing image and comes in two flavors: multi-bit schemes that encode a binary string~\cite{hidden, MBRS, fernandez2024video, trustmark, ma2022towards}, and zero-bit schemes that target detection with high statistical guarantees~\cite{BA, swift2024, latentseal}. All methods navigate the capacity--robustness--imperceptibility trade-off. Capacity is scaled either through redundant embedding~\cite{MBRS} or deeper architectures, with ChunkySeal~\cite{petrov2025hidebitsunusedwatermarking} pushing to 1024 bits where most methods cap at $\sim$100 bits. Robustness is enforced via adversarial training against standard transforms~\cite{vine}, and imperceptibility through a discriminator or a constrained watermark power targeting a specific PSNR. Standard evaluation uses bit accuracy and $\rho$-values for detection. Most methods lack adversarial training against generative attacks (with exceptions~\cite{vine}), widely considered the primary weakness. We show this concern is overstated: current generative attacks are themselves detectable thus limiting their practical threat in scenarios with high-cost detection.

\vspace{0.2cm}

\paragraph{Black-Box Attacks}
Black-box attacks assume no knowledge of the watermarking architecture and rely on query access or external oracles to guide the perturbation. Watermarks in the Sand~\cite{zhang2024watermarks} uses an image-quality oracle to iteratively refine perturbations that degrade the watermark while preserving visual fidelity. WMForger~\cite{soucek2025transferable} achieves state-of-the-art black-box removal by training a preference model via reinforcement learning to distinguish watermarked from clean images, using Fourier-domain transforms as a proxy for watermark behavior. Both methods optimize for attack success rate and distortion amplitude; neither considers whether the output is forensically distinguishable from a genuine clean image.

\vspace{0.2cm}

\paragraph{Generative Attacks}
DiffPure\cite{nie2022DiffPure} introduced diffusion-based purification by projecting the watermarked image onto the generative manifold through a forward--reverse diffusion pass. The reverse
denoising process then recover clean (purified) data, removing the watermark. CtrlRegen~\cite{liu2024ctrlregen} improves structural preservation via semantic conditioning with ControlNet. VAE-based approaches offer faster but typically weaker alternatives. 
These methods inherit artifacts of their generative backbone: diffusion-based attacks introduce characteristic frequency signatures, while VAE reconstructions suffer from smoothing \cite{corvi2023}. Overall, the removal process trades one detectable signal (the watermark) for another (the generative artifact).

\vspace{0.2cm}

\paragraph{Attack Detection}
The idea of detecting adversarial modifications has precedent in the adversarial examples literature \cite{carlini2017}, where methods such as feature squeezing~\cite{featureSqueezing} and input transformation detectors~\cite{inputTransformation} identify adversarially perturbed inputs. In watermarking, however, the analogous question, can we detect that a removal attack \emph{occurred}?, remains largely unexplored. Existing watermark robustness benchmarks~\cite{waves}
evaluate only whether the watermark survives, not whether the attack itself is detectable. We fill this gap by introducing Watermark Removal Detection as a standalone evaluation axis.

\vspace{0.2cm}

\paragraph{Image Forensics}
Detecting AI-generated images is a mature sub-field. Classifiers trained on spectral features, GAN fingerprints, or diffusion-specific artifacts~\cite{corvi2023,DIRE,AEROBLADE} achieve strong separation between real and synthetic images, though performance degrades under compression and resizing.
Our setting differs in a key respect: we do not distinguish real from AI-generated, but rather genuine clean images from images that have undergone a removal process. The forensic traces we exploit are not those of generation but of the attack process itself, a subtler signal, yet one that current attacks make no effort to quantify.

\vspace{0.2cm}

\paragraph{Image Editing Models} While generative models surged into daily usage, demand for editing models able to perform very light modifications skyrocketed. InstructPix2Pix \cite{brooks2023instructpix2pix} introduces classifier-free guidance to enable better prompt adherence, MagicBrush \cite{magic} improved on synthetic data, and subsequent models incrementally improved either image quality or diversity in the space of edits. QwenEdit \cite{wu2025qwenimagetechnicalreport} scaled the editing model up to 20B parameters. Improvements in generation enabled better editing, and recently the FLUX.2 klein \cite{flux2} model was released. The use of reinforcement learning during training enabled better prompt adherence and image quality, with FireRed \cite{firered} achieving state-of-the-art performance at low cost. Those models are famously known to remove watermark as a byproduct, thus posing threat to regulatory initiatives involving watermark \cite{swift2024}. Besides adversarial training \cite{vine} against each of these models, there is no back-up strategy for watermarker providers.

\section{Method}

Given a watermarked image $x_w$, our goal is to detect a removal attack $\mathcal{A}$ such that if the attacker wants to be sure to succeed to remove the watermark, he will eventually expose himself to the detector detailed below. 




\subsection{Detector}
\label{ssec:detector}

The forensic detector is aimed at detecting attacks, as a proxy of watermark removal. This choice stems from the property of well-designed watermarking models to hide the watermark signal $w$ from detection without the secret key~\cite{Cox200815}. A forensic detector capable of detecting the removal of a specific watermark is a security red flag, as it means that the watermarking model leaves a learnable signal, thus exposing itself to forging attacks as described in WMForger~\cite{soucek2025transferable}. 

We define our forensic detector $D_f$ as a feature extraction network with an MLP that outputs a sigmoid-based score. A score below 0.5 indicates a pristine image; above 0.5 indicates an attacked one. We train it by optimizing the Binary Cross-Entropy objective function.
We use ConvNextTiny-v2 \cite{convnextv2} as our backbone for the detector. We aim for a generalist detector since the defender has no prior knowledge of the attacker's method.
A critical design choice is separating removal detection from AI-generation detection. GenAI detectors exploit artifacts specific to a generative architecture which change as models evolve. Our detector must instead learn signatures of \textit{attacks}: the statistical traces left when part of an original image is reconstructed or edited. 
To enforce this distinction, we include WMForger in the training set alongside diffusion-based attacks. Because WMForger operates through adversarial perturbations rather than generative reconstruction, its presence forces the detector to capture broader removal artifacts. We provide more details on the training procedure in Section \ref{sec:experiment}.

\ewa{Cette allégation (WMForger presence prevents the classifier from focusing solely on generative artifacts and forces it to capture broader removal artifacts) est-elle vérifiée expérimentalement ? ie. un détecteur appris seulement sur des attaques génératives est-il moins bon ? (Sur des attaques génératives seulement bien sur) Est-ce qu'inclure WMForger dans le train set améliore les performances sur attaques génératives du test set ?}
\ewa{Addendum: en fait les résultats de la Table 4 ne vont pas en ce sens: supprimer WMForger du train set ne semble pas deteriorer les performances de detection des attaques generatives...}

\subsection{Evaluation metrics}
We evaluate watermark removal along three complementary axes: attack success rate, forensic detectability and perceptual quality.

For attack effectiveness, we measure watermark detectability and whether it leaves detectable traces. 
We evaluate the watermark survival through the lens of the corresponding $\rho$-value. We consider the null hypothesis $H_0$ that each bit of the output binary message $\hat{m}$ is independent and distributed as a Bernoulli variable with probability of success $0.5$, and the alternative hypothesis $H_1$ that $\hat{m} = m$.

We use the unified formulation from VideoSeal~\cite{fernandez2024video}, enabling comparison of $\rho$-values across different payload sizes. Let $B$ be the number of correct bits in the binary message $\hat{m}$ of length $n_{\text{bits}}$ w.r.t. $m$. Under the null hypothesis $H_0$ (random guessing), $B \sim \text{Binomial}(n_{\text{bits}}, 1/2)$. Given a message $m$ and its decoded estimate $\hat{m}$, $b = \sum_{i=1}^{n_{\text{bits}}} \mathbf{1}[\hat{m}_i = m_i]$ is the observed number of correctly decoded bits. The $\rho$-value is the probability of observing at least $b$ correct bits under $H_0$:
\begin{equation}
\begin{aligned}
\rho\text{-value}(m, \hat{m}) &= P(B \ge b \mid H_0) = \sum_{k=b}^{n_{\text{bits}}} \binom{n_{\text{bits}}}{k} \left(\frac{1}{2}\right)^{n_{\text{bits}}} \\
&= I_{1/2}\!\bigl(b,\, n_{\text{bits}} - b + 1\bigr),
\end{aligned}
\end{equation}
where $I_x(a,b)$ denotes the regularized incomplete beta function. Consequently, the Attack Success Rate (ASR) is the fraction of images whose $\rho$-value exceeds $10^{-6}$ after attack, following the standard detection threshold in the watermarking literature~\cite{stablesignature}.

\vspace{0.2cm}

To assess forensic detectability, we report the area under the ROC Curve (AUC) and the True Positive Rate at a fixed False Positive Rate (TPR@FPR) of our detector $D_f$. A high forensic detection performance signals that the attack leaves statistical artifacts. We consider false positives hamper trustworthiness of detectors and further use low FPR.

\begin{figure*}[t]
    \centering
    \includegraphics[width=\linewidth]{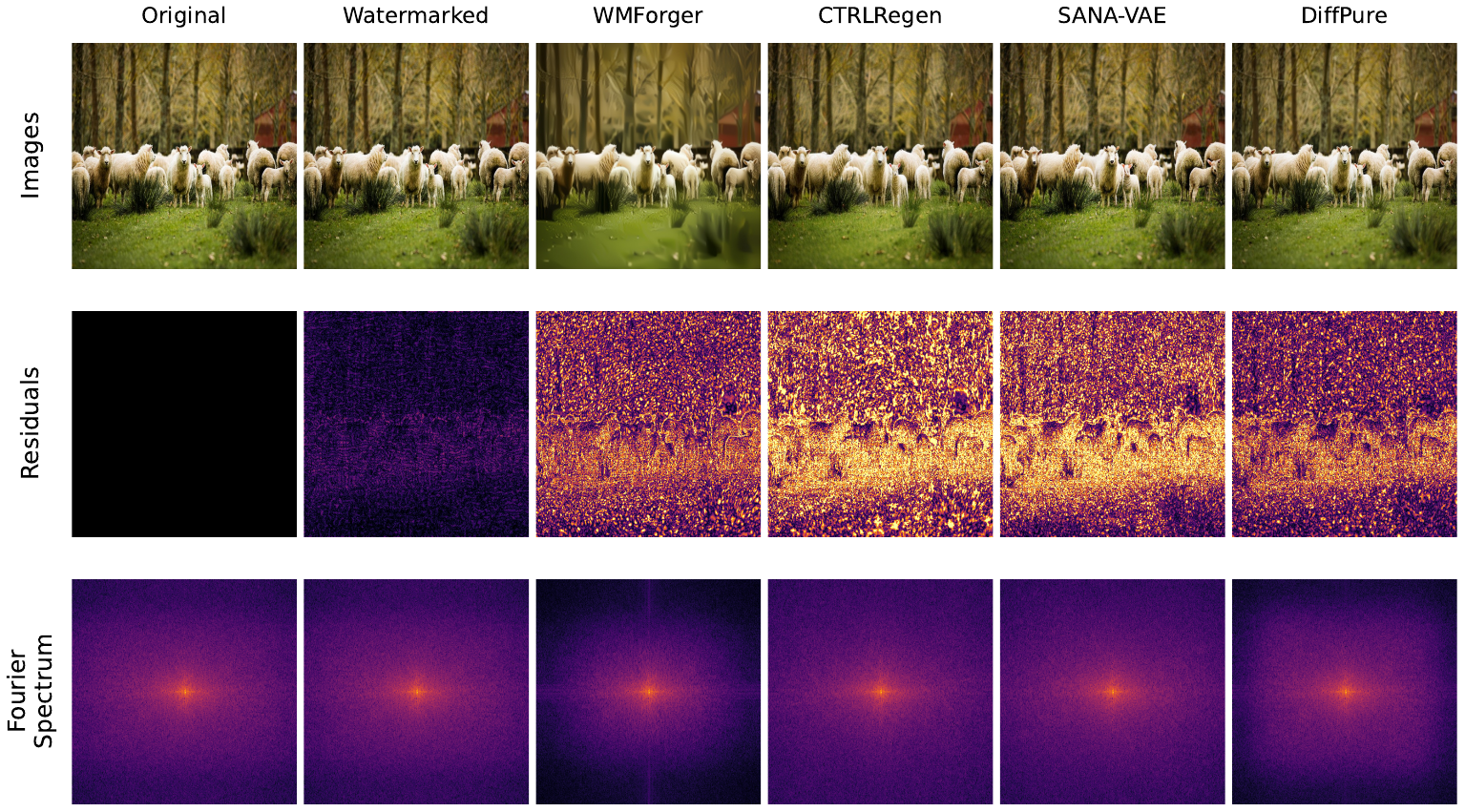}
    \caption{Watermark removal attacks samples, with residuals and Fourier spectrum. WMForger and Diffpure have the smallest residual footprint by targeting the high frequencies, which is better depicted in Fourier Spectrum figures, while CtrlRegen and SANA-VAE are spreading on the whole image their removal effort.}
    \label{fig:residuals}
\end{figure*}

\vspace{0.2cm}

For perceptual quality, we argue that commonly used pixel-level distortion metrics such as PSNR, while informative, are insufficient to fully characterize the imperceptibility of a removal attack. An attack may introduce minimal pixel-wise distortion yet shift its outputs away from the natural image manifold, producing images that are technically close to the original ones but perceptually unnatural, see Figure~\ref{fig:residuals}. To capture this distinction, we complement PSNR with LPIPS~\cite{lpips}, which measures perceptual similarity at the image level, and with the Fréchet Inception Distance (FID)~\cite{FID}, which quantifies the distributional distance between attacked images and their original, non-attacked counterparts. Together, these metrics provide a more complete picture of attack quality than distortion metrics alone.

\vspace{0.2cm}

We further focus on the attack effectiveness. To capture the joint threat in a single metric, we define the \emph{compound success rate}: the fraction of attacks that both erase the watermark \emph{and} evade forensic detection. For a set of $N$ attacked images,
\begin{equation}
\underbrace{\vphantom{(}\text{Compound}}_{\text{practical threat}} = \textcolor{asrblue}{\underbrace{\vphantom{(}\text{ASR}}_{\text{watermark erased}}} \times \textcolor{tprorange}{\underbrace{(1 - \text{TPR}@\text{FPR})}_{\text{detection evaded}}}
\label{eq:compound}
\end{equation}
where $\textcolor{asrblue}{\text{ASR}}$ is the fraction of images whose $\rho$-value exceeds $10^{-6}$ after attack, and $\textcolor{tprorange}{\text{TPR}@\text{FPR}}$ is the true positive rate of the forensic detector $D_f$ at a calibrated false positive rate. A compound rate near zero means the attack is neutralized: either the watermark survives, or the removal is detected. Because the forensics detector already relies on distortion introduced by removal attacks, we do not include perceptibility metrics in this aggregation.

\subsection{Editor Selection \& Taxonomy} To study modern image watermarking robustness against image editing, we gather a wide range of Image Editing models: InstructPix2Pix, MagicBrush,  QwenEdit, QwenEdit2511, FireRed 1.1 and FLUX 2 [klein] \cite{brooks2023instructpix2pix,magic,wu2025qwenimagetechnicalreport,firered,flux2}. This selection comprehensively illustrates the evolution of image editing and current trends in watermark robustness. We divide the study into local and global edits. A local edit targets only a specific part of the image, while a global edit affects all pixels in the scene.

\section{Experiments \& Results}
\label{sec:experiment}

\subsection{Implementation Details}

\paragraph{Dataset} We sample 5,000 images from COCO ~\cite{coco} as cover images. Each image is watermarked with two schemes, VideoSeal~\cite{fernandez2024video} and TrustMark~\cite{trustmark}, at a target PSNR of 40\,dB, yielding 10,000 watermarked images. We then apply four removal attacks to each watermarked image using hyperparameters highlighted in respective papers: DiffPure ~\cite{nie2022DiffPure} ($t=0.1$), a SANA-VAE reconstruction \cite{sana}, CtrlRegen~\cite{liu2024ctrlregen} ($\sigma=0.1$ ), and WMForger~\cite{soucek2025transferable} (500 steps, lr = 0.05). Including pristine originals and watermarked images as negatives, the full dataset comprises approximately 50k images, split 70/10/20 by original image ID to prevent leakage between train, validation, and test sets. The dataset is composed of one quarter pristine images (original and watermarked) and three quarters attacked images.

\vspace{0.2cm}

\paragraph{Watermarking methods} We use the public implementations of VideoSeal, 1.0 version at 256 bits, and TrustMark model. Those two models are heavily used and VideoSeal is considered as state-of-the-art. Both are post-hoc multi-bit watermarking schemes. We watermark at $256 \times 256$ resolution. The choice of two architecturally distinct schemes ensures our detector results don't rely on one watermarker, but rather on the attacks performed. We do not use adversarially trained watermark models such as Vine \cite{vine} or EditGuard \cite{zhang2024editguard} as they trade robustness to editing against capacity or perceptibility. We chose to sidestep this road and design a defense in an orthogonal fashion with respect to the watermark training procedure.

\vspace{0.2cm}

\paragraph{Detector training} We finetune a ConvNeXtV2-Tiny~\cite{convnextv2} pretrained on ImageNet-22k \cite{deng2009imagenet}, replacing the classification head with a two-layer MLP (hidden dim 128, GELU activation, dropout 0.33). All images are resized to $224 \times 224$. We train for 8 epochs on a single NVIDIA L40S with batch size 256, using AdamW with weight decay $2.5 \times 10^{-3}$, and a cosine schedule with one-epoch linear warmup. The classifier head uses learning rate $4.8 \times 10^{-4}$ while backbone parameters are finetuned at $0.32\times$ that rate. To improve robustness against editorial post-processing, we apply random augmentations during training: JPEG compression (quality 60--100), Gaussian blur (radius 0.5--2.0), downscale-upscale (scale 0.5--1.2), and Gaussian noise ($\sigma$ 0.01--0.08). Each image has a 20\% probability of receiving one randomly selected augmentation.

\begin{figure}[t]
    \centering
    \includegraphics[width=1\linewidth]{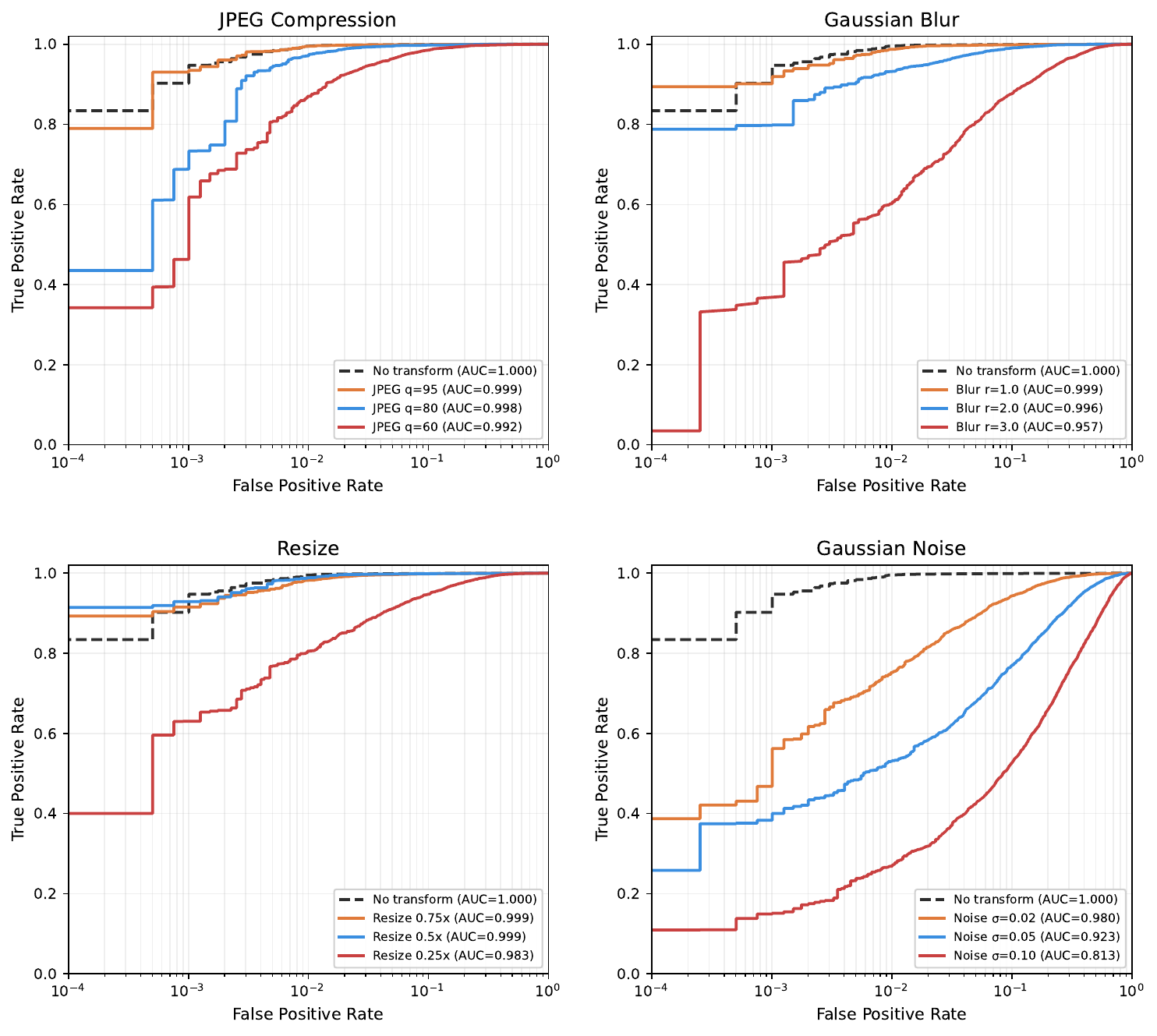}
    \caption{Detector robustness under post-processing. ROC curves (log-scale FPR) for four transform families at increasing strength. The detector maintains AUC $>$ 0.988 under JPEG, blur, and resize. Gaussian noise is the best evasion strategy but requires levels visibly degrading the image. \ewa{AUC=1 pour No transform? Faut regarder combien de chiffres après la virgule pour avoir la vraie valeur ?} \ewa{On doit pouvoir retrouver les valeurs de la Table 1 (averaged sur toutes les attaques) sur les courbes No transform}}
    \label{fig:robustness_roc}
\end{figure}

\subsection{Attack effectiveness vs.\ perceptual quality}
Perceptibility metrics and robustness results are reported in Table~\ref{tab:quality_metrics} and Table~\ref{tab:watermark_survival}, respectively. Across all metrics, removal attacks cause substantially more perturbations than the initial watermarking process. Without specific knowledge of the watermark's spatial distribution or strength, attacks must apply brute-force, global alterations, ultimately resulting in poor perceptibility scores. 

Hence, robustness and perceptibility results reveal a clear tradeoff between attack success and image quality. Looking at PSNR alone, WMForger and DiffPure appear as the most performative methods. However, LPIPS and FID tell a different story: WMForger produces images that deviate significantly from the natural image manifold, with a FID of 30, confirming that PSNR on its own is insufficient as it only accounts for perturbation amplitude and not distributional proximity to natural images. When factoring in robustness, we observe a negative correlation between ASR and perceptual quality: DiffPure based on Flux.1~[dev] achieves the best FID and LPIPS, but the worst ASR on VideoSeal; similarly, SANA-VAE obtains the best FID and second-best LPIPS while underperforming on robustness. Conversely, the most aggressive watermark removal techniques degrade image quality the most, highlighting the trade-off between removal and imperceptibility. More qualitative samples are given in Figure~\ref{fig:residuals}.

\begin{table}[t]
\centering
\caption{Image quality metrics between original and variant images. PSNR ($\uparrow$), LPIPS ($\downarrow$), and FID ($\downarrow$).}
\label{tab:quality_metrics}
\resizebox{\linewidth}{!}{%
\begin{tabular}{ll ccc}
\toprule
Watermarker & Image Version & PSNR $\uparrow$ & LPIPS $\downarrow$ & FID $\downarrow$ \\
\midrule
\multirow{1}{*}{None}
& Watermarked       & $40.1 \pm 1.9$ & $0.004 \pm 0.020$ & 1.13 \\
\midrule

\multirow{4}{*}{TrustMark}

 & DiffPure ($t{=}0.1$)    & $29.6 \pm 3.4$ & $0.068 \pm 0.035$ & 4.01 \\
 & SANA-VAE          & $24.8 \pm 4.2$ & $0.087 \pm 0.040$ & 3.82 \\
 & WMForger          & $29.5 \pm 2.7$ & $0.216 \pm 0.079$ & 30.42 \\
 & CtrlRegen ($\sigma{=}0.1$) & $24.1 \pm 3.4$ & $0.115 \pm 0.043$ & 5.34 \\
\midrule
\multirow{5}{*}{VideoSeal}
 & DiffPure ($t{=}0.1$)    & $29.6 \pm 3.4$ & $0.067 \pm 0.034$ & 3.76 \\
 & SANA-VAE          & $24.8 \pm 4.1$ & $0.089 \pm 0.038$ & 3.84 \\
 & WMForger          & $29.5 \pm 2.7$ & $0.215 \pm 0.079$ & 30.14 \\
 & CtrlRegen ($\sigma{=}0.1$) & $24.1 \pm 3.4$ & $0.115 \pm 0.042$ & 5.23 \\
\bottomrule
\end{tabular}%
}
\end{table}

\begin{table}[t]
\centering
\caption{Watermark survival and forensic detectability. ASR
per attack, TPR at FPR=$10^{-3}$, and compound success rate
ASR~$\times$~(1$-$TPR@$10^{-3}$). Bold marks highest compound
per watermarker.}
\label{tab:watermark_survival}
\renewcommand{\arraystretch}{1.2}
\setlength{\tabcolsep}{4pt}
\begin{tabular}{llccc}
\toprule
WM & Attack & ASR & TPR@$10^{-3}$ & Compound \\
\midrule
\multirow{4}{*}{\textbf{TM}}
 & DiffPure    & 0.997 & 0.785 & \textbf{0.214} \\
 & SANA-VAE    & 0.797 & 0.944 & 0.045 \\
 & WMForger    & 1.000 & 0.998 & 0.002 \\
 & CtrlRegen   & 1.000 & 0.936 & 0.064 \\
\midrule
\multirow{4}{*}{\textbf{VS}}
 & DiffPure    & 0.569 & 0.810 & \textbf{0.108} \\
 & SANA-VAE    & 1.000 & 0.927 & 0.073 \\
 & WMForger    & 1.000 & 1.000 & 0.000 \\
 & CtrlRegen   & 1.000 & 0.915 & 0.085 \\
\bottomrule
\end{tabular}
\end{table}

\subsection{Detection Performance}

Table~\ref{tab:operative_points} reports the performance of our forensic detector at two decision thresholds, calibrated on a held-out set of 10k COCO images (distinct from other sets) to achieve False Positive Rates of $10^{-2}$ and $10^{-3}$. At FPR=$10^{-2}$, the detector identifies the majority of attacked images across both watermarking schemes, with TPR exceeding 86\% for all attacks except DiffPure based on Flux.1~[dev] (67--68\%). WMForger is the easiest to detect, reaching 99.8\% TPR for both TrustMark and VideoSeal. At the stricter FPR=$10^{-3}$ operating point, WMForger remains highly detectable (80--81\%), while SANA-VAE and CtrlRegen degrade to 34--43\%. DiffPure drops to roughly 25\%, making it the most evasive method. These results mirror the removal-imperceptibility tradeoff observed earlier: WMForger, despite achieving high ASR, leaves the strongest forensic traces and worst perceptual quality, while DiffPure trades removal performance for both better image fidelity and greater evasiveness.  

To emphasize the tradeoff, we use the compound success rate defined in Equation (\ref{eq:compound}), which measures the fraction of attacks that
both remove the watermark and evade forensic detection. Results in Table~\ref{tab:watermark_survival} reveal
that high ASR alone is insufficient to characterize a practical
threat. The best attack overall is Diffpure, and achieve a compound success rate of 10\%, making even the best attack unsuitable in our setting.

\subsection{Cross-watermark generalization} A potential concern is that the detector learns artifacts specific to the watermarker rather than to the removal attack itself. While this should not happen with carefully crafted watermark models, WMForger \cite{soucek2025transferable} shows some modern watermarking models are not content-aware, thus reproducing the same artifacts from one image to another, allowing for a straight-forward forging of the watermark. 
To investigate this, we train three additional variants at matched data quantity ($\sim$20k training images each): each disjointedly trained and tested on Videoseal, TrustMark or non-watermarked images (Table~\ref{tab:cross_wm}). AUC remains stable across all conditions ($\approx$~0.989-0.995), confirming that the detector captures attack artifacts that transfer across architecturally distinct watermarking schemes. Point estimates of TPR at FPR~$10^{-3}$ vary between conditions (0.701--0.615) which is not significative. The full model (TPR~=~0.920, trained on $\sim$41k images) substantially outperforms all matched conditions, confirming that training set size is the dominant factor at strict operating points. We conclude that the detector is not watermarker-dependent: cross-watermarker transfer is robust, and the remaining gap to the full model is explained by data quantity, not watermarker diversity.

\begin{table}[t]
\centering
\caption{Watermark-agnosticism of the detector. All conditions use
$\sim$20k training images. Performance is stable whether the detector
is trained on TrustMark, VideoSeal, or no watermarked images at all,
indicating it learns attack artifacts rather than watermark-specific signals.}
\label{tab:cross_wm}
\renewcommand{\arraystretch}{1.2}
\setlength{\tabcolsep}{4pt}
\begin{tabular}{lccc}
\toprule
Watermarker & AUC & TPR@$10^{-2}$ & TPR@$10^{-3}$ \\
\midrule
TrustMark  & 0.995 & 0.922 & 0.701 \\
VideoSeal  & 0.994 & 0.900 & 0.615 \\
None       & 0.989 & 0.845 & 0.618 \\
\bottomrule
\end{tabular}
\end{table}

\begin{table}[t]
\centering
\caption{Watermark Removal Detection. Detection rates at operating
points FPR=$10^{-2}$ and $10^{-3}$ (calibrated on COCO).
$\pm$ denotes bootstrap standard deviation over test set resamples.}
\label{tab:operative_points}
\renewcommand{\arraystretch}{1.2}
\begin{tabular}{llcc}
\toprule
& & \multicolumn{2}{c}{TPR@FPR} \\
\cmidrule(l){3-4}
Watermarker & Method & $10^{-2}$ & $10^{-3}$ \\
\midrule
\multirow{4}{*}{TrustMark}
 & DiffPure ($t{=}0.1$) \cite{flux2024} & 0.974\ci{0.014} & 0.785\ci{0.095} \\
 & VAE (SANA) \cite{sana}               & 0.999\ci{0.002} & 0.944\ci{0.042} \\
 & WMForger \cite{soucek2025transferable}& 1.000\ci{0.001} & 0.998\ci{0.001} \\
 & CtrlRegen \cite{liu2024ctrlregen}    & 0.996\ci{0.004} & 0.936\ci{0.042} \\
\midrule
\multirow{4}{*}{VideoSeal}
 & DiffPure                             & 0.987\ci{0.008} & 0.810\ci{0.096} \\
 & SANA VAE                             & 0.998\ci{0.002} & 0.927\ci{0.053} \\
 & WMForger                             & 1.000\ci{0.000} & 1.000\ci{0.001} \\
 & CtrlRegen                            & 0.997\ci{0.003} & 0.915\ci{0.050} \\
\midrule
\multirow{1}{*}{All}
 & All attacks                          & 0.994\ci{0.003} & 0.920\ci{0.037} \\
\bottomrule
\end{tabular}
\end{table}

\subsection{Robustness to Post-Processing}

A rational adversary might apply standard image transforms after removal to wash away forensic traces. We evaluate the detector under four families of common post-processing at increasing severity: JPEG compression (quality 95, 80, 60), Gaussian blur (radius 1.0, 2.0, 3.0), downscale-upscale (factors 0.75, 0.5, 0.25), and additive Gaussian noise ($\sigma =$ 0.02, 0.05, 0.10). Figure~\ref{fig:robustness_roc} reports ROC curves for each family.

The detector is highly resilient to JPEG, blur, and resize: AUC remains above 0.99 across all two-first values for these three families. Gaussian noise is the most effective countermeasure, degrading AUC to 0.813 at $\sigma = 0.10$. However, noise at this level introduces visible degradation (PSNR $\approx$ 20\,dB), defeating the adversary's goal of preserving image quality. At the more realistic $\sigma = 0.02$, the detector still achieves AUC 0.980. 

\subsection{Generalization to Unseen Attacks}
To assess zero-shot robustness, we conducted leave-one-attack-out
experiments in Table~\ref{tab:loo}: we use the same training procedure
without a specific attack in the train and validation sets. The results
reveal an expected generalization gap: the average AUC drops from a
near-perfect $0.998$ on seen attacks to $0.902$ on the held-out set.
Notably, the detector generalizes better to WMForger and CtrlRegen
(both $\approx 0.98$), which are the removal methods that leave the
strongest forensic traces and degrade perceptual quality the most.
In contrast, detection rates are lower for cleaner methods such as
DiffPure and SANA-VAE ($13.3\%$ and $36.5\%$), whose subtler artifacts
are harder to pick up without explicit training. This confirms that our
forensic approach is most effective precisely where it matters most:
attacks that successfully remove the watermark tend to do so at the
cost of heavier image distortion, which in turn makes them easier to detect.

\begin{table}[t]
\centering
\caption{Leave-one-attack-out generalization results for the forensic detector on COCO test split.}
\label{tab:loo}
\resizebox{\linewidth}{!}{
\begin{tabular}{lccc}
\toprule
Held-out attack & Other attacks AUC & Held-out attack AUC & Gap \\
\midrule
DiffPure                      & 0.9998 & 0.7670 & +0.2329 \\
SANA-VAE                      & 0.9974 & 0.8682 & +0.1292 \\
WMForger                      & 0.9970 & 0.9876 & +0.0094 \\
CtrlRegen ($\sigma=0.1$)      & 0.9974 & 0.9845 & +0.0130 \\
\midrule
Average                       & 0.9979 & 0.9018 & +0.0961 \\
\bottomrule
\end{tabular}
}
\end{table}

\begin{table}[t]
\centering
\caption{AEROBLADE baseline vs.\ our forensic detector in the leave-one-out setting (average AUC on COCO test split, per attack).}
\label{tab:aeroblade-vs-ours}
\resizebox{\linewidth}{!}{
\begin{tabular}{lccccc}
\toprule
Method & DiffPure & SANA-VAE & WMForger & CtrlRegen & Average \\
\midrule
AEROBLADE       & 0.5835 & 0.5245 & 0.8685 & 0.5871 & 0.6409 \\
Ours (held-out) & \textbf{0.7670} &\textbf{ 0.8682} & \textbf{0.9876} & \textbf{0.9845} & \textbf{0.9018} \\
\bottomrule
\end{tabular}
}
\end{table}

\subsection{Detection of GenAI}
\label{sec:genai_detection}
We evaluate the detector on 1000 FLUX-1 [dev] generations, using COCO captions as prompt for each image. Using the decision threshold 0.9999 calibrated on COCO natural images, we achieve FPR at $10^{-3}$ with only 1 image flagged as attacked out of 1000.
This supports our claim of detecting attacks and not solely GenAI artifacts as planned by our training procedure in Section \ref{ssec:detector}.
The detector is not picking up generic GenAI artifacts.

The converse also holds: \textit{a GenAI detector is not a watermark-removal detector}. We evaluate AEROBLADE~\cite{AEROBLADE}, which scores images by minimum reconstruction error across a set of VAEs, on our removal attacks (Table~\ref{tab:aeroblade-vs-ours}). In our setting, we observe it fails largely, even for CTRLRegen which leverages a classic "stabilityai/sd-vae-ft-mse" VAE. For fair comparison, our forensic detector performs better in its leave-one-out setting with an average AUC of $0.9$ against AEROBLADE average at $0.64$. Removal artifacts do not place images on the generated-image manifold; they leave a distinct signature that GenAI detectors miss.

\subsection{Comparison with Adversarial Training}

Since watermark removal detection is unexplored, no apple-to-apple comparison with an existing baseline is possible. The closest alternative is adversarial training of the watermarker against known removal methods \cite{vine}. Table~\ref{tab:defense_comparison} contrasts the two strategies. The approaches are complementary rather than competing: adversarial training strengthens the watermark signal, while our detector catches attacks that break through. Crucially, our framework can be deployed on top of any existing watermarking scheme without retraining it, at the cost of a single lightweight classifier.

\begin{table}[t]
\centering
\caption{Comparison of defense strategies against watermark removal.}
\label{tab:defense_comparison}
\resizebox{\linewidth}{!}{
\begin{tabular}{lcc}
\toprule
 & Adversarial Training & WRD (Ours) \\
\midrule
No watermarker retraining & \ding{55} & \ding{51} \\
Preserves original quality--capacity tradeoff & \ding{55} & \ding{51} \\
Deployable on closed-source schemes & \ding{55} & \ding{51} \\
Generalizes to heavy attacks\textsuperscript{*} & \ding{55} & \ding{51} \\
Defense mechanism & Watermark robustness & Post-hoc detection \\
\bottomrule

\end{tabular}
}
\end{table}

\subsection{Image Editing}

Image editing models are increasingly proposed as a practical attack vector against watermarking. We challenge this proposition under the three-axis evaluation, and find they face the same fundamental dilemma.
We evaluate eight image editors spanning 4 years of development.
\textbf{InstructPix2Pix}~\cite{brooks2023instructpix2pix} (2022) and \textbf{MagicBrush}~\cite{magic} (2023) are finetuned Stable Diffusion models, both run with 50 denoising steps and an image guidance scale of~1.5.
\textbf{Qwen-Image-Edit}~\cite{wu2025qwenimagetechnicalreport} (2024) is a flow-matching diffusion model run at 50 steps with a classifier-free guidance scale of~4.0.
\textbf{Qwen-2511}~\cite{wu2025qwenimagetechnicalreport} (2025) uses an updated checkpoint of the same architecture at 40 steps.
\textbf{FireRed~1.1}~\cite{firered} (2026) builds on the same backbone with a true CFG scale of~4.0 at 50 steps; its rapid variant substitutes a distilled transformer and runs in only 4 steps with CFG disabled.
\textbf{FLUX 2 [klein]}~\cite{flux2} (2025) is a 9B-parameter rectified-flow transformer run at 4 steps; its KV variant incorporates key-value caching for efficient inference.
All editors except InstructPix2Pix and MagicBrush operate in bfloat16 precision on images resized to $512\times512$ pixels, with fixed seeds for reproducibility.

\vspace{0.4cm}

\begin{figure*}
    \centering
    \includegraphics[width=0.9\linewidth]{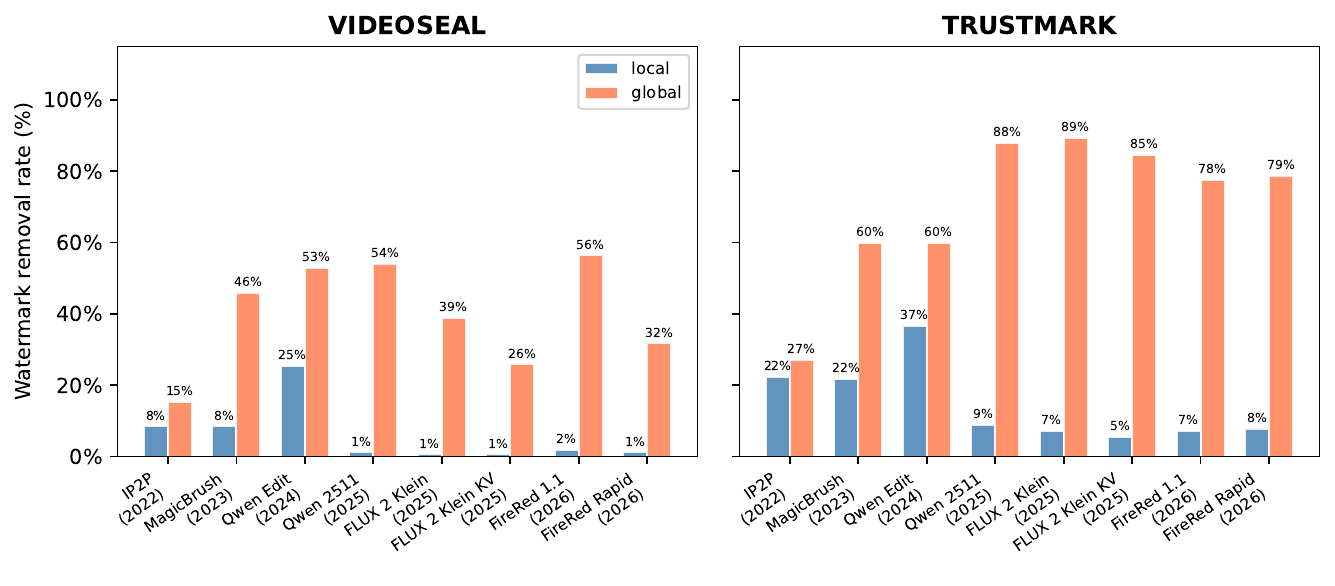}
    \caption{Watermark removal rate per editor. After a peak in 2024, watermarking removal began to lessen as a byproduct of Image Editing.  }
    \label{fig:editing_removal} 
\end{figure*}

\paragraph{Watermark Survival vs.\ Editing Quality}
Fig.~\ref{fig:editing_removal} investigates how watermark survival rate evolved throughout the years, broken down by edit type. Following the EMU Edit taxonomy, we classify edits tagged as \texttt{style}, \texttt{global}, or \texttt{background} as global edits since they are designed to affect the entire image, while all remaining categories are treated as local edits. InstructPix2Pix barely disrupted the watermark, as most edits were not performed due to overly precise, out-of-domain prompts. MagicBrush improved prompt adherence and disrupted the watermark more aggressively. Qwen-Image-Edit continued this trend, until its late-2025 version which began optimizing for image quality preservation alongside prompt adherence. FLUX~2 Klein and FireRed confirmed this trend, achieving strong editing performance without compromising image fidelity, thus preserving the watermark. Notably, global edits consistently yield higher removal rates than local ones, as they alter low-level statistics across the entire image. An interesting finding is that watermark removal drops significantly for local editing, making it less of a threat. 

\begin{figure*}
    
    \centering
    \includegraphics[width=1\linewidth]{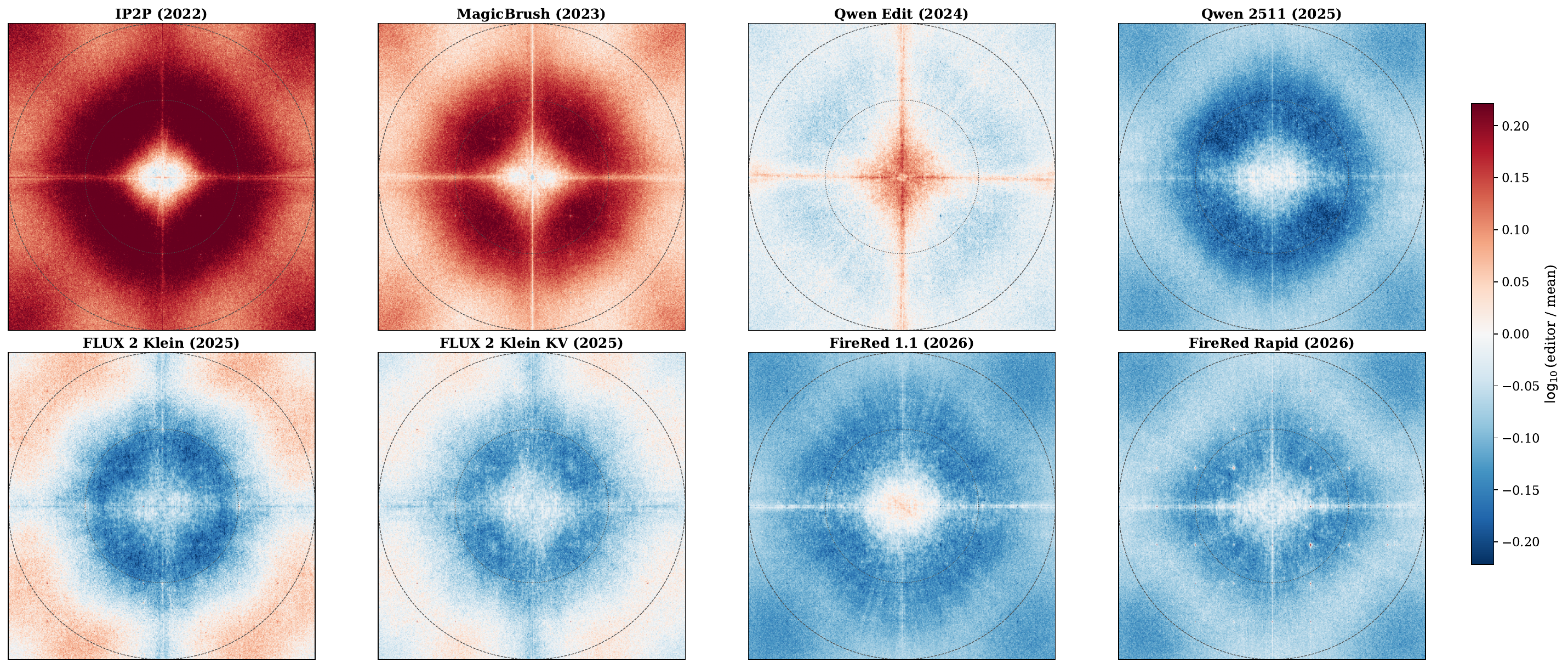}
\caption{Mean log-ratio of the 2D Fourier spectrum of each editor's perturbation relative to the average across all editors (local edits only). Red indicates frequencies more strongly affected by a given editor than average; blue indicates the opposite. The spectral
footprint of image editing has shifted from low frequencies (IP2P,
MagicBrush) to mid-to-high frequencies (Qwen 2511, FLUX~2 Klein, FireRed),
reflecting a trend toward finer-grained edits that preserve
low-level image statistics and with them the watermark signal.}
\label{fig:spectral_analysis}
\end{figure*}

At first glance, it appears that \textit{better editors make for safer watermarks}.
We further study this trend through the lens of spectral analysis: in Figure~\ref{fig:spectral_analysis}, we compute the Fourier spectrum of the difference between each watermarked image and its edited counterpart over 250 samples per editor, yielding a per-editor frequency fingerprint that sheds light on what transformations occur in the image domain. We observe a global trend in the frequency domain where modern editors change much lightly the image, especially in the high-frequency part of the spectrum. Table~\ref{tab:perceptual} depicts perceptual metrics computed between edited images and their input (watermarked) counterparts. Those results support the trend of improving editing models in term of perceptibility. 

\begin{table}[ht]
\centering
\caption{Perceptual quality of edited images.}
\label{tab:perceptual}
\setlength{\tabcolsep}{0.5pt}
\begin{tabular}{lcccccc}
\toprule
 & \multicolumn{3}{c}{Local edits} & \multicolumn{3}{c}{Global edits} \\
\cmidrule(lr){2-4} \cmidrule(lr){5-7}
Editor & PSNR$\uparrow$ & SSIM$\uparrow$ & LPIPS$\downarrow$ & PSNR$\uparrow$ & SSIM$\uparrow$ & LPIPS$\downarrow$ \\
\midrule
IP2P (2022)        & 18.4 & 0.646 & 0.287 & 15.50 & 0.573 & 0.335 \\
MagicBrush (2023)  & 21.3 & 0.720 & 0.173 & 12.51 & 0.459 & 0.470 \\
Qwen Edit (2024)   & 20.5 & 0.744 & 0.223 & 14.61 & 0.529 & 0.428 \\
Qwen 2511 (2025)   & 22.2 & 0.882 & 0.142 & 10.25 & 0.360 & 0.575 \\
FLUX 2 Klein (2025)    & 23.0 & 0.888 & 0.104 & 10.19 & 0.345 & 0.575 \\
FLUX 2 Klein KV (2025) & 22.5 & 0.878 & 0.115 & 11.16 & 0.394 & 0.520 \\
FireRed 1.1 (2026)     & 21.0 & 0.876 & 0.145 & 10.69 & 0.413 & 0.511 \\
FireRed Rapid (2026)   & 22.0 & 0.875 & 0.154 & 13.05 & 0.500 & 0.454 \\
\bottomrule
\end{tabular}
\end{table}

\begin{figure*}
    
    \centering
    \includegraphics[width=1\linewidth]{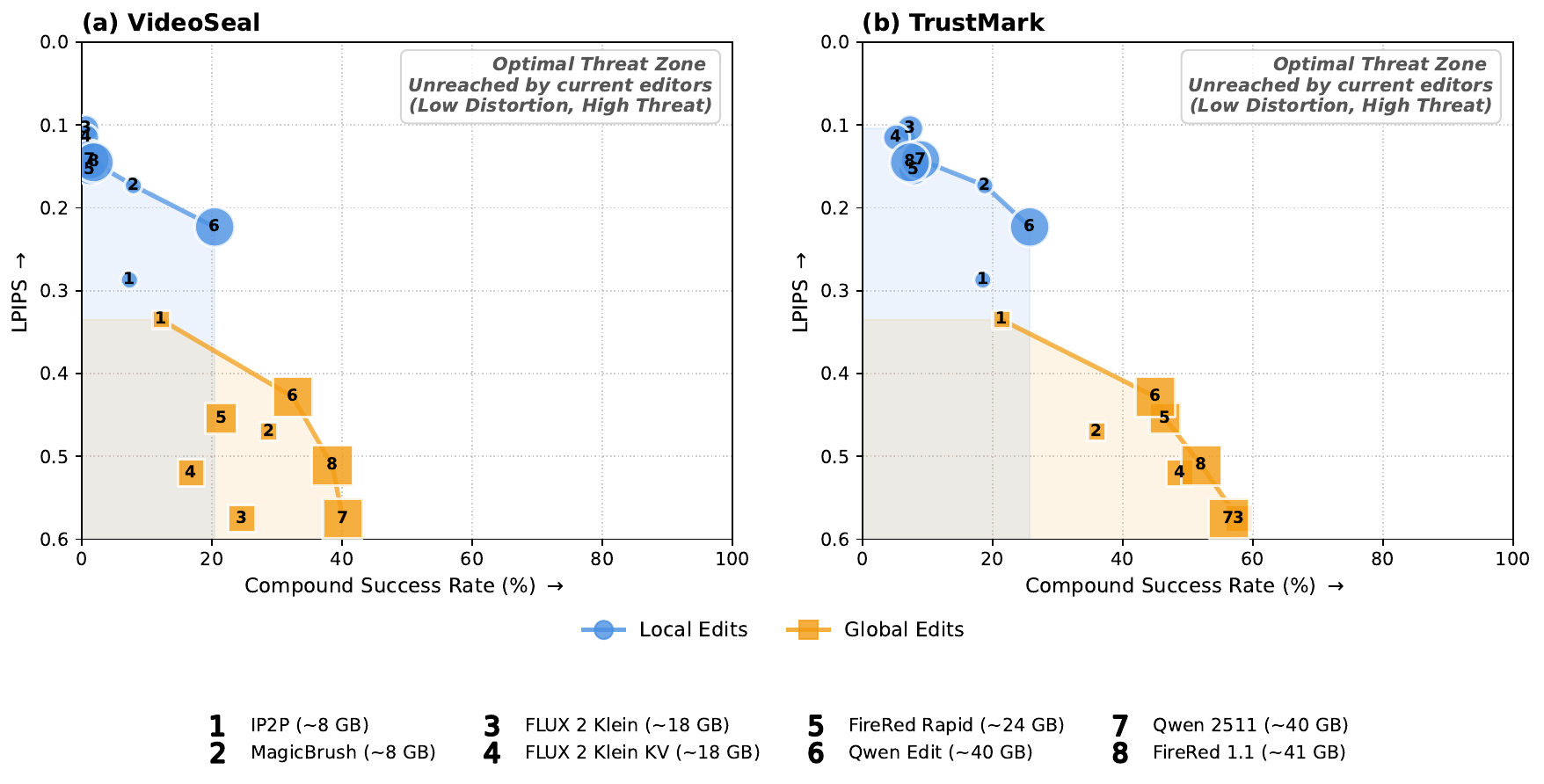}
\caption{LPIPS and compound success rate performance of Image Editing Models. We use (a) Videoseal and (b) TrustMark as watermarking models. Across a wide range of model size (from 8GB to 41GB), no model is able to reach a satisfactory compound success rate at a decent LPIPS.} 
\label{fig:pareto}
\end{figure*}

\vspace{0.2cm}

\paragraph{WRD Performance on Edited Images}
\label{sssec:wrd_editing}
As computed in Table~\ref{tab:perceptual}, PSNR is rather low: at best 23.0 db for local edits against 15.5 db for global ones.  Unlike watermark removal attacks that reach at worst 24 db of PSNR, and mostly around 0.100 of LPIPS, those very specific image transforms alter significantly the input image. Given this observation, we make the following hypothesis: \textit{Image Editing is fatally flawed to bypass watermarking} as it is either not removing the watermark, thus failing as an attack, or it is disrupting enough to leave detectable forensic signals. To validate this hypothesis, we test our forensic detector in a zero-shot fashion on 250 edited images for each editor. We provide averaged results on different settings: how do the detector perform if the watermark survived or was the removed (Table~\ref{tab:forensic_detection_all_editors}), and additionnaly under local and global edits at different level of FPR (Table~\ref{tab:forensic_editors_local_global}), using previously computed thresholds on the held-out calibration set from COCOVal2017. Average watermark survival rate is reported in the "WM survived column". What stands out is $D_f$ still achieves very high TPR in a zero-shot setting, suggesting Image Editing makes up for a highly disruptive transform, easy to detect, even at very low FPR. Best Image Editing models in term of quality and inference speed, such as FireRed Rapid and Flux 2 [klein] kv, are close to being detected half the time even in this stringent setting. Figure \ref{fig:pareto} emphasizes even more the practical state of  "image editing models as removal attacks". The frontier is far away from the optimal threat zone that guarantees to the attacker to both remove the watermark and evade the forensic detection, whether the edit is global or local. Table \ref{tab:compound_editors} details compound success rates.

\begin{table}[t]
\centering
\caption{Zero-shot compound success rate (\%) per editor at FPR${}=10^{-3}$: $\textcolor{asrblue}{\text{ASR}} \times \textcolor{tprorange}{(1 - \text{TPR})}$. Higher means a more practical threat.}
\label{tab:compound_editors}
\setlength{\tabcolsep}{4pt}
\begin{tabular}{l cc cc}
\toprule
& \multicolumn{2}{c}{\textbf{VideoSeal}} & \multicolumn{2}{c}{\textbf{TrustMark}} \\
\cmidrule(lr){2-3} \cmidrule(lr){4-5}
& Local & Global & Local & Global \\
\midrule
IP2P (2022)              & 7.3 & 12.2 & 18.5 & 21.4 \\
MagicBrush (2023)        & 7.9 & 28.7 & 18.8 & 36.0 \\
Qwen Edit (2024)         & 20.4 & 32.4 & 25.7 & 45.0 \\
Qwen 2511 (2025)         & 1.2 & 40.1 &  8.9 & 56.3 \\
FLUX 2 [klein] (2025)    & 0.6 & 24.6 &  7.3 & 57.8 \\
FLUX 2 [klein] KV (2025) & 0.6 & 16.8 &  5.2 & 48.8 \\
FireRed 1.1 (2026)       & 1.8 & 38.5 &  7.3 & 52.1 \\
FireRed Rapid (2026)     & 1.2 & 21.4 &  7.9 & 46.5 \\
\bottomrule
\end{tabular}
\end{table}

\vspace{0.2cm}

\paragraph{Temporal trend}
A clear temporal trend emerges from Figure~\ref{fig:editing_removal}: watermark removal as a byproduct of image editing peaked around 2024 with QwenEdit, reaching removal rates of 53-60\% depending on the watermarker (Tables~\ref{tab:forensic_detection_all_editors} and~\ref{tab:forensic_editors_local_global}), and has since declined sharply with more recent models. FLUX~2 Klein and FireRed, despite representing the current state of the art in editing quality, remove the watermark in fewer than 10\% of local edits (Table~\ref{tab:forensic_editors_local_global}). This trend is not coincidental. Figure~\ref{fig:spectral_analysis} reveals its spectral correlate: early editors (IP2P, MagicBrush) applied broad low-frequency perturbations that inadvertently disrupted the watermark signal, while modern editors concentrate their changes in mid-to-high frequencies, preserving the low-level image statistics that carry the watermark. The shift was driven by demand for higher image fidelity, not by any watermark-awareness on the part of model developers, yet it incidentally makes watermarks less threatened by this attack vector. Importantly, this means image editing is not the ubiquitous threat described in literature \cite{vine}: as Section~\ref{sssec:wrd_editing} shows, the editors that do succeed in removing the watermark introduce distortions strong enough to be forensically detected, placing them in the same fundamental dilemma as dedicated removal attacks.

\begin{table*}[t]
\centering
\caption{Watermark survival and Zero-shot Forensic TPR rates (\%), conditioned on whether the watermark survived or was destroyed by the edit. Results shown for target FPR $\in \{10^{-2}, 10^{-3}, 10^{-4}\}$. Bold $\geq 50\%$.}
\label{tab:forensic_detection_all_editors}
\setlength{\tabcolsep}{2pt}
\begin{tabular}{l c cc cc cc c cc cc cc}
\toprule
& \multicolumn{7}{c}{\textbf{VideoSeal}} & \multicolumn{7}{c}{\textbf{TrustMark}} \\
\cmidrule(lr){2-8} \cmidrule(lr){9-15}
& \makecell{WM\\Survived} & \multicolumn{2}{c}{\makecell{TPR\\(FPR${}=10^{-2}$)}} & \multicolumn{2}{c}{\makecell{TPR\\(FPR${}=10^{-3}$)}} & \multicolumn{2}{c}{\makecell{TPR\\(FPR${}=10^{-4}$)}} & \makecell{WM\\Survived} & \multicolumn{2}{c}{\makecell{TPR\\(FPR${}=10^{-2}$)}} & \multicolumn{2}{c}{\makecell{TPR\\(FPR${}=10^{-3}$)}} & \multicolumn{2}{c}{\makecell{TPR\\(FPR${}=10^{-4}$)}} \\
\cmidrule(lr){3-4} \cmidrule(lr){5-6} \cmidrule(lr){7-8} \cmidrule(lr){10-11} \cmidrule(lr){12-13} \cmidrule(lr){14-15}
&  & \scriptsize WM intact & \scriptsize WM lost & \scriptsize WM intact & \scriptsize WM lost & \scriptsize WM intact & \scriptsize WM lost &  & \scriptsize WM intact & \scriptsize WM lost & \scriptsize WM intact & \scriptsize WM lost & \scriptsize WM intact & \scriptsize WM lost \\
\midrule
IP2P & 89.2 & 13.0 & \textbf{77.8} & 9.4 & \textbf{70.4} & 7.6 & \textbf{55.6} & 76.0 & 12.6 & \textbf{58.3} & 8.4 & \textbf{51.7} & 6.8 & 46.7 \\
MagicBrush & 78.8 & 8.6 & \textbf{69.8} & 7.1 & \textbf{56.6} & 4.6 & \textbf{50.9} & 65.2 & 19.6 & \textbf{50.6} & 12.9 & 41.4 & 8.6 & 35.6 \\
Qwen Edit & 65.2 & 3.7 & \textbf{73.6} & 3.1 & \textbf{70.1} & 0.0 & \textbf{62.1} & 54.0 & 0.0 & \textbf{69.6} & 0.0 & \textbf{60.9} & 0.0 & 43.5 \\
Qwen 2511 & 80.8 & 2.0 & 47.9 & 1.5 & 39.6 & 0.5 & 29.2 & 62.9 & 0.0 & 44.2 & 0.0 & 34.6 & 0.0 & 25.0 \\
FLUX 2 [klein] & 86.4 & 9.7 & \textbf{52.9} & 7.4 & 44.1 & 5.6 & 29.4 & 64.8 & 0.6 & 43.2 & 0.6 & 34.1 & 0.6 & 26.1 \\
FLUX 2 [klein] KV & 90.8 & 12.3 & \textbf{73.9} & 7.5 & \textbf{65.2} & 5.3 & 47.8 & 67.6 & 6.5 & \textbf{58.0} & 4.7 & 44.4 & 3.6 & 28.4 \\
FireRed 1.1 & 79.6 & 4.5 & \textbf{51.0} & 3.5 & 39.2 & 2.5 & 23.5 & 68.8 & 2.9 & 39.7 & 1.7 & 33.3 & 1.2 & 20.5 \\
FireRed Rapid & 88.4 & 6.8 & \textbf{72.4} & 5.0 & \textbf{62.1} & 3.6 & \textbf{55.2} & 68.0 & 5.9 & 38.8 & 3.5 & 27.5 & 2.4 & 25.0 \\
\bottomrule
\end{tabular}
\end{table*}

\begin{table*}[t]
\centering
\caption{Watermark survival and Zero-shot Forensic TPR rates (\%) per editor, broken down by edit type (local vs.\ global). WM Survived reports the watermark survival rate. Results shown for target FPR $\in \{10^{-2}, 10^{-3}, 10^{-4}\}$. Bold $\geq 50\%$.}
\label{tab:forensic_editors_local_global}
\setlength{\tabcolsep}{3pt}
\begin{tabular}{l cc cc cc cc cc cc cc cc}
\toprule
& \multicolumn{8}{c}{\textbf{VideoSeal}} & \multicolumn{8}{c}{\textbf{TrustMark}} \\
\cmidrule(lr){2-9} \cmidrule(lr){10-17}
& \multicolumn{2}{c}{\makecell{WM\\Survived}} & \multicolumn{2}{c}{\makecell{TPR\\(FPR${}=10^{-2}$)}} & \multicolumn{2}{c}{\makecell{TPR\\(FPR${}=10^{-3}$)}} & \multicolumn{2}{c}{\makecell{TPR\\(FPR${}=10^{-4}$)}} & \multicolumn{2}{c}{\makecell{WM\\Survived}} & \multicolumn{2}{c}{\makecell{TPR\\(FPR${}=10^{-2}$)}} & \multicolumn{2}{c}{\makecell{TPR\\(FPR${}=10^{-3}$)}} & \multicolumn{2}{c}{\makecell{TPR\\(FPR${}=10^{-4}$)}} \\
\cmidrule(lr){2-3} \cmidrule(lr){4-5} \cmidrule(lr){6-7} \cmidrule(lr){8-9} \cmidrule(lr){10-11} \cmidrule(lr){12-13} \cmidrule(lr){14-15} \cmidrule(lr){16-17}
& \scriptsize Local & \scriptsize Global & \scriptsize Local & \scriptsize Global & \scriptsize Local & \scriptsize Global & \scriptsize Local & \scriptsize Global & \scriptsize Local & \scriptsize Global & \scriptsize Local & \scriptsize Global & \scriptsize Local & \scriptsize Global & \scriptsize Local & \scriptsize Global \\
\midrule
IP2P (2022) & 91.5 & 84.7 & 18.2 & 23.5 & 13.9 & 20.0 & 12.1 & 14.1 & 77.6 & 72.9 & 21.8 & 27.1 & 17.6 & 21.2 & 17.0 & 15.3 \\
MagicBrush (2023) & 91.5 & 54.1 & 8.5 & 47.1 & 7.3 & 37.6 & 4.8 & 32.9 & 78.2 & 40.0 & 18.8 & \textbf{52.9} & 13.9 & 40.0 & 8.5 & 36.5 \\
Qwen Edit (2024) & 74.5 & 47.1 & 21.2 & 41.2 & 20.0 & 38.8 & 16.4 & 31.8 & 63.3 & 40.0 & 30.0 & 35.0 & 30.0 & 25.0 & 23.3 & 15.0 \\
Qwen 2511 (2025) & 98.8 & 45.9 & 0.0 & 31.8 & 0.0 & 25.9 & 0.0 & 17.6 & 91.1 & 12.0 & 0.0 & 46.0 & 0.0 & 36.0 & 0.0 & 26.0 \\
FLUX 2 [klein] (2025) & 99.4 & 61.2 & 0.0 & 45.9 & 0.0 & 36.5 & 0.0 & 25.9 & 92.7 & 10.6 & 0.6 & 44.7 & 0.6 & 35.3 & 0.6 & 27.1 \\
FLUX 2 [klein] KV (2025) & 99.4 & 74.1 & 1.8 & 49.4 & 1.2 & 35.3 & 1.2 & 24.7 & 94.5 & 15.3 & 6.7 & \textbf{55.3} & 4.8 & 42.4 & 2.4 & 29.4 \\
FireRed 1.1 (2026) & 98.2 & 43.5 & 0.6 & 40.0 & 0.0 & 31.8 & 0.0 & 20.0 & 92.7 & 22.4 & 1.2 & 40.0 & 0.6 & 32.9 & 0.0 & 21.2 \\
FireRed Rapid (2026) & 98.8 & 68.2 & 1.8 & 38.8 & 0.6 & 32.9 & 0.0 & 28.2 & 92.1 & 21.2 & 2.4 & 43.5 & 0.6 & 31.8 & 0.0 & 28.2 \\
\bottomrule
\end{tabular}
\end{table*}

\section{Limitations}

Our approach has three main limitations:

\paragraph{Zero-shot detection of subtle attacks.} While the detector easily generalizes to artifact-heavy removal methods (e.g., WMForger, CtrlRegen), zero-shot performance drops against cleaner generative attacks like DiffPure, see Table~\ref{tab:loo}. However, this issue is mitigated in practice: these evasive attacks are computationally expensive and do not guarantee successful watermark removal (Table~\ref{tab:watermark_survival}), reducing their viability for a rational attacker in a high-cost detection scenario.

\vspace{0.2cm}

\paragraph{Vulnerability to adaptive attacks.} Like any learned model, our detector is susceptible to adversarial evasion. If the model is made public, even in a black-box manner, an adaptive adversary could use gradient-free methods (e.g., SurFree~\cite{surfree}) to craft perturbations that bypass detection. Practical deployment therefore requires keeping the model weights private and restricting detection access to maximize the cost of such attacks. 

\vspace{0.2cm}

\paragraph{Experimental scope.} Our current evaluation focuses on post-hoc watermarks at a $256 \times 256$ resolution, with cover images being $512 \times 512$. Future work should investigate whether removal artifacts remain as easily detectable on generation-time watermarks (e.g., Stable Signature) or on high-resolution images, where forensic traces may be highly localized.

\section{Discussion}

\paragraph{What is a good enough detector?} Detection performance must be interpreted relative to the deployment scenario. At $10^{-3}$ FPR, a platform processing one million images per day would wrongly flag 1,000 clean images. Whether this is acceptable depends on the base rate of detection and the cost of false accusations. In a targeted monitoring setting, screening high-profile accounts for top-down misinformation, the volume is low which makes our detector directly applicable. In a mass-screening setting, the detector is better suited as a triage step that flags candidates for secondary review rather than as a standalone verdict.

\vspace{0.2cm}

\paragraph{The removal/evasion trade-off.} Our robustness experiments reveal that an effective removal strategy requires to visibly degrade the image. This creates a fundamental dilemma for the adversary: to let the watermark or the removal attack be detected. Current removal methods offer no path through this trade-off, which we view as the central practical implication of our work.

\section{Conclusion}
We introduced Watermark Removal Detection as a missing evaluation axis, and our results reframe the practical threat posed by watermark removal.
The compound success rate reveals that even the best attack (DiffPure) succeeds only 10\% of the time once forensic detection is accounted for, while methods with perfect ASR such as WMForger are effectively neutralized.
Additional work on Image Editing models uncovers them as being optimize with objectives favoring watermark survival and still being detectable by the forensic detector. 
The practical implication is immediate: platforms can deploy a forensic detector alongside the watermark detector, creating a two-check system deployable on any existing watermarking scheme without retraining, at the cost of a single lightweight classifier.

Our work has clear limitations established in existing literature. The detector is vulnerable to adaptive adversaries who optimize against it, thus prompting to keep the detector private, and performance degrades on out-of-distribution images and attacks.

Until an attack can pass both the watermark detector and the forensic detector, watermark removal remains an incomplete threat.
\bibliographystyle{IEEEtran}
\bibliography{references}

@String{Computer = "{IEEE} Computer" }

@String{Springer = "Springer-Verlag" }

@inproceedings{trustmark,
  title = {TrustMark: Robust Watermarking and Watermark Removal for Arbitrary Resolution Images},
  author={Bui, Tu and Agarwal, Shruti and Collomosse, John},
  booktitle = {IEEE International Conference on Computer Vision (ICCV)},
  year = {2025},
  month = oct
}

@article{fernandez2024video,
  title={Video Seal: Open and Efficient Video Watermarking},
  author={Fernandez, Pierre and Elsahar, Hady and Yalniz, I. Zeki and Mourachko, Alexandre},
  journal={arXiv preprint arXiv:2412.09492},
  year={2024}
}

@inproceedings{swift2024,
  title={SWIFT: Semantic Watermarking for Image Forgery Thwarting},
  author={Gautier Evennou and Vivien Chappelier and Ewa Kijak and Teddy Furon},
  booktitle={Proceedings of the IEEE International Workshop on Information Forensics and Security (WIFS)},
  year={2024}
}

@inproceedings{hidden,
author = {Zhu, Jiren and Kaplan, Russell and Johnson, Justin and Fei-Fei, Li},
title = {HiDDeN: Hiding Data With Deep Networks},
year = {2018},
isbn = {978-3-030-01266-3},
publisher = {Springer-Verlag},
address = {Berlin, Heidelberg},
url = {https://doi.org/10.1007/978-3-030-01267-0_40},
doi = {10.1007/978-3-030-01267-0_40},
booktitle = {Computer Vision – ECCV 2018: 15th European Conference, Munich, Germany, September 8-14, 2018, Proceedings, Part XV},
pages = {682–697},
numpages = {16},
location = {Munich, Germany}
}

@article{liu2024ctrlregen,
  title={Image watermarks are removable using controllable regeneration from clean noise},
  author={Liu, Yepeng and Song, Yiren and Ci, Hai and Zhang, Yu and Wang, Haofan and Shou, Mike Zheng and Bu, Yuheng},
  journal={arXiv preprint arXiv:2410.05470},
  year={2024}
}

@inproceedings{MBRS,
author = {Jia, Zhaoyang and Fang, Han and Zhang, Weiming},
year = {2021},
month = {10},
pages = {41-49},
title = {MBRS: Enhancing Robustness of DNN-based Watermarking by Mini-Batch of Real and Simulated JPEG Compression},
doi = {10.1145/3474085.3475324}
}

@inproceedings{nie2022DiffPure,
  title={Diffusion Models for Adversarial Purification},
  author={Nie, Weili and Guo, Brandon and Huang, Yujia and Xiao, Chaowei and Vahdat, Arash and Anandkumar, Anima},
  booktitle = {International Conference on Machine Learning (ICML)},
  year={2022}
}

@inproceedings{soucek2025transferable,
  title={Transferable Black-Box One-Shot Forging of Watermarks via Image Preference Models},
  author={Sou\v{c}ek, Tom\'{a}\v{s} and Rebuffi, Sylvestre-Alvise and Fernandez, Pierre and Jovanovi\'{c}, Nikola and Elsahar, Hady and Lacatusu, Valeriu and Tran, Tuan and Mourachko, Alexandre},
  booktitle={Advances in Neural Information Processing Systems},
  year={2025}
}

@misc{latentseal,
      title={Fast, Secure, and High-Capacity Image Watermarking with Autoencoded Text Vectors}, 
      author={Gautier Evennou and Vivien Chappelier and Ewa Kijak},
      year={2025},
      eprint={2510.00799},
      archivePrefix={arXiv},
      primaryClass={cs.CR},
      url={https://arxiv.org/abs/2510.00799}, 
}

@inproceedings{ma2022towards, title={Towards Blind Watermarking: Combining Invertible and Non-invertible Mechanisms}, author={Ma, Rui and Guo, Mengxi and Hou, Yi and Yang, Fan and Li, Yuan and Jia, Huizhu and Xie, Xiaodong}, booktitle={Proceedings of the 30th ACM International Conference on Multimedia}, pages={1532--1542}, year={2022} }

@inproceedings{vine,
  author       = {Shilin Lu and
                  Zihan Zhou and
                  Jiayou Lu and
                  Yuanzhi Zhu and
                  Adams Wai{-}Kin Kong},
  title        = {Robust Watermarking Using Generative Priors Against Image Editing:
                  From Benchmarking to Advances},
  booktitle    = {The Thirteenth International Conference on Learning Representations,
                  {ICLR} 2025, Singapore, April 24-28, 2025},
  publisher    = {OpenReview.net},
  year         = {2025},
  url          = {https://openreview.net/forum?id=16O8GCm8Wn},
  bibsource    = {dblp computer science bibliography, https://dblp.org}
}

@inproceedings{coco,
  author       = {Tsung{-}Yi Lin and
                  Michael Maire and
                  Serge J. Belongie and
                  James Hays and
                  Pietro Perona and
                  Deva Ramanan and
                  Piotr Doll{\'{a}}r and
                  C. Lawrence Zitnick},
  editor       = {David J. Fleet and
                  Tom{\'{a}}s Pajdla and
                  Bernt Schiele and
                  Tinne Tuytelaars},
  title        = {Microsoft {COCO:} Common Objects in Context},
  booktitle    = {Computer Vision - {ECCV} 2014 - 13th European Conference, Zurich,
                  Switzerland, September 6-12, 2014, Proceedings, Part {V}},
  series       = {Lecture Notes in Computer Science},
  volume       = {8693},
  pages        = {740--755},
  publisher    = {Springer},
  year         = {2014},
  url          = {https://doi.org/10.1007/978-3-319-10602-1\_48},
  doi          = {10.1007/978-3-319-10602-1\_48},
}

@inproceedings{convnextv2,
  author       = {Sanghyun Woo and
                  Shoubhik Debnath and
                  Ronghang Hu and
                  Xinlei Chen and
                  Zhuang Liu and
                  In So Kweon and
                  Saining Xie},
  title        = {ConvNeXt {V2:} Co-designing and Scaling ConvNets with Masked Autoencoders},
  booktitle    = {{IEEE/CVF} Conference on Computer Vision and Pattern Recognition,
                  {CVPR} 2023, Vancouver, BC, Canada, June 17-24, 2023},
  pages        = {16133--16142},
  publisher    = {{IEEE}},
  year         = {2023},
  url          = {https://doi.org/10.1109/CVPR52729.2023.01548},
  doi          = {10.1109/CVPR52729.2023.01548},

}

@inproceedings{
      zhang2024watermarks,
      title={Watermarks in the Sand: Impossibility of Strong Watermarking for Generative Models},
      author={Hanlin Zhang and Benjamin L. Edelman and Danilo Francati and Daniele Venturi and Giuseppe Ateniese and Boaz Barak},
      booktitle={Forty-first International Conference on Machine Learning},
      year={2024},
}

@article{BA,
  TITLE = {{Broken Arrows}},
  AUTHOR = {Furon, Teddy and Bas, Patrick},
  JOURNAL = {{EURASIP Journal on Information Security}},
  PUBLISHER = {{Hindawi/SpringerOpen}},
  VOLUME = {2008},
  PAGES = {ID 597040},
}

@misc{petrov2025hidebitsunusedwatermarking,
      title={We Can Hide More Bits: The Unused Watermarking Capacity in Theory and in Practice}, 
      author={Aleksandar Petrov and Pierre Fernandez and Tomáš Souček and Hady Elsahar},
      year={2025},
      eprint={2510.12812},
      archivePrefix={arXiv},
      primaryClass={cs.CR},
      url={https://arxiv.org/abs/2510.12812}, 
}

@inproceedings{surfree,
  author       = {Thibault Maho and
                  Teddy Furon and
                  Erwan Le Merrer},
  title        = {SurFree: {A} Fast Surrogate-Free Black-Box Attack},
  booktitle    = {{IEEE} Conference on Computer Vision and Pattern Recognition, {CVPR}
                  2021, virtual, June 19-25, 2021},
  pages        = {10430--10439},
  publisher    = {Computer Vision Foundation / {IEEE}},
  year         = {2021},
  doi          = {10.1109/CVPR46437.2021.01029},
}

@inproceedings{waves,
author = {An, Bang and Ding, Mucong and Rabbani, Tahseen and Agrawal, Aakriti and Xu, Yuancheng and Deng, Chenghao and Zhu, Sicheng and Mohamed, Abdirisak and Wen, Yuxin and Goldstein, Tom and Huang, Furong},
title = {WAVES: benchmarking the robustness of image watermarks},
year = {2024},
booktitle = {Proceedings of the 41st International Conference on Machine Learning},
series = {ICML'24}
}

@INPROCEEDINGS{DIRE,
  author={Wang, Zhendong and Bao, Jianmin and Zhou, Wengang and Wang, Weilun and Hu, Hezhen and Chen, Hong and Li, Houqiang},
  booktitle={2023 IEEE/CVF International Conference on Computer Vision (ICCV)}, 
  title={DIRE for Diffusion-Generated Image Detection}, 
  year={2023},
  volume={},
  number={},
  pages={22388-22398},
  doi={10.1109/ICCV51070.2023.02051}}

@INPROCEEDINGS {AEROBLADE,
author = { Ricker, Jonas and Lukovnikov, Denis and Fischer, Asja },
booktitle = { 2024 IEEE/CVF Conference on Computer Vision and Pattern Recognition (CVPR) },
title = {{ AEROBLADE: Training-Free Detection of Latent Diffusion Images Using Autoencoder Reconstruction Error }},
year = {2024},
doi = {10.1109/CVPR52733.2024.00872},
publisher = {IEEE Computer Society},
address = {Los Alamitos, CA, USA},
}

@INPROCEEDINGS{corvi2023,
  author={Corvi, Riccardo and Cozzolino, Davide and Zingarini, Giada and Poggi, Giovanni and Nagano, Koki and Verdoliva, Luisa},
  booktitle={IEEE International Conference on Acoustics, Speech and Signal Processing (ICASSP)}, 
  title={On The Detection of Synthetic Images Generated by Diffusion Models}, 
  year={2023}
}

@inproceedings{carlini2017,
author = {Carlini, Nicholas and Wagner, David},
title = {Adversarial Examples Are Not Easily Detected: Bypassing Ten Detection Methods},
year = {2017},
booktitle = {Proceedings of the 10th ACM Workshop on Artificial Intelligence and Security},
pages = {3–14},
numpages = {12}
}

@inproceedings{featureSqueezing,
  author       = {Weilin Xu and
                  David Evans and
                  Yanjun Qi},
  title        = {Feature Squeezing: Detecting Adversarial Examples in Deep Neural Networks},
  booktitle    = {25th Annual Network and Distributed System Security Symposium, {NDSS}
                  2018, San Diego, California, USA, February 18-21, 2018},
  year         = {2018},
}

@ARTICLE{inputTransformation,
  author={Nesti, Federico and Biondi, Alessandro and Buttazzo, Giorgio},
  journal={IEEE Transactions on Neural Networks and Learning Systems}, 
  title={Detecting Adversarial Examples by Input Transformations, Defense Perturbations, and Voting}, 
  year={2023},
  doi={10.1109/TNNLS.2021.3105238}}

@inproceedings{sana,
  author       = {Enze Xie and
                  Junsong Chen and
                  Junyu Chen and
                  Han Cai and
                  Haotian Tang and
                  Yujun Lin and
                  Zhekai Zhang and
                  Muyang Li and
                  Ligeng Zhu and
                  Yao Lu and
                  Song Han},
  title        = {{SANA:} Efficient High-Resolution Text-to-Image Synthesis with Linear
                  Diffusion Transformers},
  booktitle    = {The Thirteenth International Conference on Learning Representations,
                  {ICLR} 2025, Singapore, April 24-28, 2025},
  year         = {2025},
}

@inproceedings{deng2009imagenet,
  title={Imagenet: A large-scale hierarchical image database},
  author={Deng, Jia and Dong, Wei and Socher, Richard and Li, Li-Jia and Li, Kai and Fei-Fei, Li},
  booktitle={2009 IEEE conference on computer vision and pattern recognition},
  pages={248--255},
  year={2009},
  organization={IEEE}
}

@misc{flux2024,
    author={Black Forest Labs},
    title={FLUX},
    year={2024},
    howpublished={\url{https://github.com/black-forest-labs/flux}},
}

@inproceedings{FID,
  author       = {Martin Heusel and
                  Hubert Ramsauer and
                  Thomas Unterthiner and
                  Bernhard Nessler and
                  Sepp Hochreiter},
  title        = {GANs Trained by a Two Time-Scale Update Rule Converge to a Local Nash
                  Equilibrium},
  booktitle    = {Advances in Neural Information Processing Systems 30: Annual Conference
                  on Neural Information Processing Systems 2017, December 4-9, 2017,
                  Long Beach, CA, {USA}},
  year         = {2017},
}

@inproceedings{lpips,
  author       = {Richard Zhang and
                  Phillip Isola and
                  Alexei A. Efros and
                  Eli Shechtman and
                  Oliver Wang},
  title        = {The Unreasonable Effectiveness of Deep Features as a Perceptual Metric},
  booktitle    = {2018 {IEEE} Conference on Computer Vision and Pattern Recognition,
                  {CVPR} 2018, Salt Lake City, UT, USA, June 18-22, 2018},
}

@misc{castro2025watermarking,
  author       = {Castro, Wes and Yalniz, Zeki},
  title        = {Video Invisible Watermarking at Scale},
  year         = {2025},
  month        = nov,
  howpublished = {Engineering at Meta},
  url          = {https://engineering.fb.com/2025/11/04/video-engineering/video-invisible-watermarking-at-scale/},
  note         = {Accessed: 2026-02-20}
}

@legislation{euaiact2024,
  author       = {{European Parliament and Council of the European Union}},
  title        = {Regulation ({EU}) 2024/1689 of the {European Parliament} and of the {Council} of 13 {June} 2024 laying down harmonised rules on artificial intelligence ({Artificial Intelligence Act})},
  year         = {2024},
  journal      = {Official Journal of the European Union},
  volume       = {L 2024/1689},
  date         = {2024-07-12},
  url          = {https://eur-lex.europa.eu/eli/reg/2024/1689/oj}
}

@inproceedings{stablesignature,
  author       = {Pierre Fernandez and
                  Guillaume Couairon and
                  Herv{\'{e}} J{\'{e}}gou and
                  Matthijs Douze and
                  Teddy Furon},
  title        = {The Stable Signature: Rooting Watermarks in Latent Diffusion Models},
  booktitle    = {{IEEE/CVF} International Conference on Computer Vision, {ICCV} 2023,
                  Paris, France, October 1-6, 2023},
  publisher    = {{IEEE}},
  year         = {2023},
  url          = {https://doi.org/10.1109/ICCV51070.2023.02053},
  doi          = {10.1109/ICCV51070.2023.02053},
}

@inproceedings{treerings,
 author = {Wen, Yuxin and Kirchenbauer, John and Geiping, Jonas and Goldstein, Tom},
 booktitle = {Advances in Neural Information Processing Systems},
 editor = {A. Oh and T. Naumann and A. Globerson and K. Saenko and M. Hardt and S. Levine},
 pages = {58047--58063},
 publisher = {Curran Associates, Inc.},
 title = {Tree-Rings Watermarks: Invisible Fingerprints for Diffusion Images},
 url = {https://proceedings.neurips.cc/paper_files/paper/2023/file/b54d1757c190ba20dbc4f9e4a2f54149-Paper-Conference.pdf},
 volume = {36},
 year = {2023}
}

@incollection{Cox200815,
title = {Chapter 2 - Applications and Properties},
editor = {Ingemar J. Cox and Matthew L. Miller and Jeffrey A. Bloom and Jessica Fridrich and Ton Kalker},
booktitle = {Digital Watermarking and Steganography (Second Edition)},
publisher = {Morgan Kaufmann},
edition = {Second Edition},
address = {Burlington},
pages = {15-59},
year = {2008},
series = {The Morgan Kaufmann Series in Multimedia Information and Systems},
isbn = {978-0-12-372585-1},
doi = {https://doi.org/10.1016/B978-012372585-1.50005-X},
url = {https://www.sciencedirect.com/science/article/pii/B978012372585150005X},
author = {Ingemar J. Cox and Matthew L. Miller and Jeffrey A. Bloom and Jessica Fridrich and Ton Kalker}
}

@misc{firered,
      title={FireRed-Image-Edit-1.0 Technical Report}, 
      author={Super Intelligence Team and Changhao Qiao and Chao Hui and Chen Li and Cunzheng Wang and Dejia Song and Jiale Zhang and Jing Li and Qiang Xiang and Runqi Wang and Shuang Sun and Wei Zhu and Xu Tang and Yao Hu and Yibo Chen and Yuhao Huang and Yuxuan Duan and Zhiyi Chen and Ziyuan Guo},
      year={2026},
      eprint={2602.13344},
      archivePrefix={arXiv},
      primaryClass={cs.CV},
      url={https://arxiv.org/abs/2602.13344}, 
}

@misc{flux2,
    author={Black Forest Labs},
    title={{FLUX.2: Frontier Visual Intelligence}},
    year={2025},
    howpublished={\url{https://bfl.ai/blog/flux-2}},
}

@misc{wu2025qwenimagetechnicalreport,
      title={Qwen-Image Technical Report}, 
      author={Chenfei Wu and Jiahao Li and Jingren Zhou and Junyang Lin and Kaiyuan Gao and Kun Yan and Sheng-ming Yin and Shuai Bai and Xiao Xu and Yilei Chen and Yuxiang Chen and Zecheng Tang and Zekai Zhang and Zhengyi Wang and An Yang and Bowen Yu and Chen Cheng and Dayiheng Liu and Deqing Li and Hang Zhang and Hao Meng and Hu Wei and Jingyuan Ni and Kai Chen and Kuan Cao and Liang Peng and Lin Qu and Minggang Wu and Peng Wang and Shuting Yu and Tingkun Wen and Wensen Feng and Xiaoxiao Xu and Yi Wang and Yichang Zhang and Yongqiang Zhu and Yujia Wu and Yuxuan Cai and Zenan Liu},
      year={2025},
      eprint={2508.02324},
      archivePrefix={arXiv},
      primaryClass={cs.CV},
      url={https://arxiv.org/abs/2508.02324}, 
}

@inproceedings{magic,
    title={MagicBrush: A Manually Annotated Dataset for Instruction-Guided Image Editing},
    author={Kai Zhang and Lingbo Mo and Wenhu Chen and Huan Sun and Yu Su},
    booktitle={Thirty-seventh Conference on Neural Information Processing Systems Datasets and Benchmarks Track},
    year={2023},

}

@INPROCEEDINGS{brooks2023instructpix2pix,
  author={Brooks, Tim and Holynski, Aleksander and Efros, Alexei A.},
  booktitle={2023 IEEE/CVF Conference on Computer Vision and Pattern Recognition (CVPR)}, 
  title={InstructPix2Pix: Learning to Follow Image Editing Instructions}, 
  year={2023},
  volume={},
  number={},
  pages={18392-18402},
  doi={10.1109/CVPR52729.2023.01764}}

@inproceedings{zhang2024editguard,
  title={Editguard: Versatile image watermarking for tamper localization and copyright protection},
  author={Zhang, Xuanyu and Li, Runyi and Yu, Jiwen and Xu, Youmin and Li, Weiqi and Zhang, Jian},
  booktitle={Proceedings of the IEEE/CVF Conference on Computer Vision and Pattern Recognition},
  pages={11964--11974},
  year={2024}
}

@misc{evennou2026forensiccostwatermarkremoval,
      title={The Forensic Cost of Watermark Removal}, 
      author={Gautier Evennou and Ewa Kijak},
      year={2026},
      eprint={2604.25491},
      archivePrefix={arXiv},
      primaryClass={cs.CV},
      url={https://arxiv.org/abs/2604.25491}, 
}
\end{document}